\documentclass{article} 
\usepackage{iclr2024_conference,times}

\usepackage{amsmath,amsfonts,bm}

\def\eqref#1{equation~\ref{#1}}

\def\1{\bm{1}}

\DeclareMathAlphabet{\mathsfit}{\encodingdefault}{\sfdefault}{m}{sl}
\SetMathAlphabet{\mathsfit}{bold}{\encodingdefault}{\sfdefault}{bx}{n}

\usepackage{hyperref}
\usepackage{url}
\usepackage{amsmath}
\usepackage{graphicx}
\usepackage{booktabs}
\usepackage{cleveref}
\usepackage[utf8]{inputenc}

\title{GigaEvo: An Open Source Optimization Framework Powered by LLMs and Evolution Algorithms}

\author{Valentin Khrulkov\thanks{Artificial Intelligence Research Institute (AIRI), Presnenskaya Embankment 6, bld. 2, Moscow, 123112, Russia. Corresponding author: \texttt{khrulkov.v@gmail.com}} \And
Andrey V. Galichin$^{*}$ \And
Denis Bashkirov\thanks{Sber, 19 Vavilova St., Moscow 117312, Russia} \And
Dmitry Vinichenko$^{\dagger}$ 
\AND
Oleg Travkin$^{\dagger}$ \And
Roman Alferov$^{\dagger}$ \And
Andrey Kuznetsov$^{*}$ \And
Ivan Oseledets$^{*}$
}
\iclrfinalcopy 
\begin{document}

\maketitle

\begin{abstract}
Recent advances in LLM-guided evolutionary computation, particularly \textit{AlphaEvolve}~\citep{novikov2025alphaevolve,georgiev2025mathematical}, have demonstrated remarkable success in discovering novel mathematical constructions and solving challenging optimization problems.
However, the high-level descriptions in published work leave many implementation details unspecified, hindering reproducibility and further research.
In this report we present \textit{GigaEvo}, an extensible open-source framework that enables researchers to study and experiment with hybrid LLM-evolution approaches inspired by \textit{AlphaEvolve}.
Our system provides modular implementations of key components: MAP-Elites quality-diversity algorithms, asynchronous DAG-based evaluation pipelines, LLM-driven mutation operators with insight generation and bidirectional lineage tracking, and flexible multi-island evolutionary strategies.
In order to assess reproducibility and validate our implementation we evaluate GigaEvo on challenging problems from the \textit{AlphaEvolve} paper: Heilbronn triangle placement, circle packing in squares, and high-dimensional kissing numbers.
The framework emphasizes modularity, concurrency, and ease of experimentation, enabling rapid prototyping through declarative configuration.
We provide detailed descriptions of system architecture, implementation decisions, and experimental methodology to support further research in LLM-driven evolutionary methods. The GigaEvo framework and all experimental code are available at \href{https://github.com/AIRI-Institute/gigaevo-core}{https://github.com/AIRI-Institute/gigaevo-core}.
\end{abstract}

\section{Introduction}
The recent paper \citep{novikov2025alphaevolve} introduced \textit{AlphaEvolve}, 
a framework that combines large language model (LLM) code generation with evolutionary computation, 
achieving state-of-the-art results on challenging algorithmic and mathematical problems. 
The paper's description is necessarily high-level, leaving many design and implementation choices unspecified.

In this report, we present \textit{GigaEvo}, an extensible open-source implementation designed to enable reproducibility studies and further research. 
Our system emphasizes modularity, concurrency, and ease of experimentation, providing researchers with a foundation for exploring hybrid LLM-evolution approaches. 
We evaluate the framework on benchmark problems from \citep{novikov2025alphaevolve} to assess whether published results can be reproduced with alternative implementation strategies and to validate our architectural choices. Additionally, we test our framework on practical tasks across various domains. The GigaEvo framework and all experimental code are available at \href{https://github.com/FusionBrainLab/gigaevo-core}{https://github.com/FusionBrainLab/gigaevo-core}.

\section{System Design}
\paragraph{Redis Database.}
\begin{figure}[htb!]
    \centering
    \includegraphics[width=0.95\linewidth]{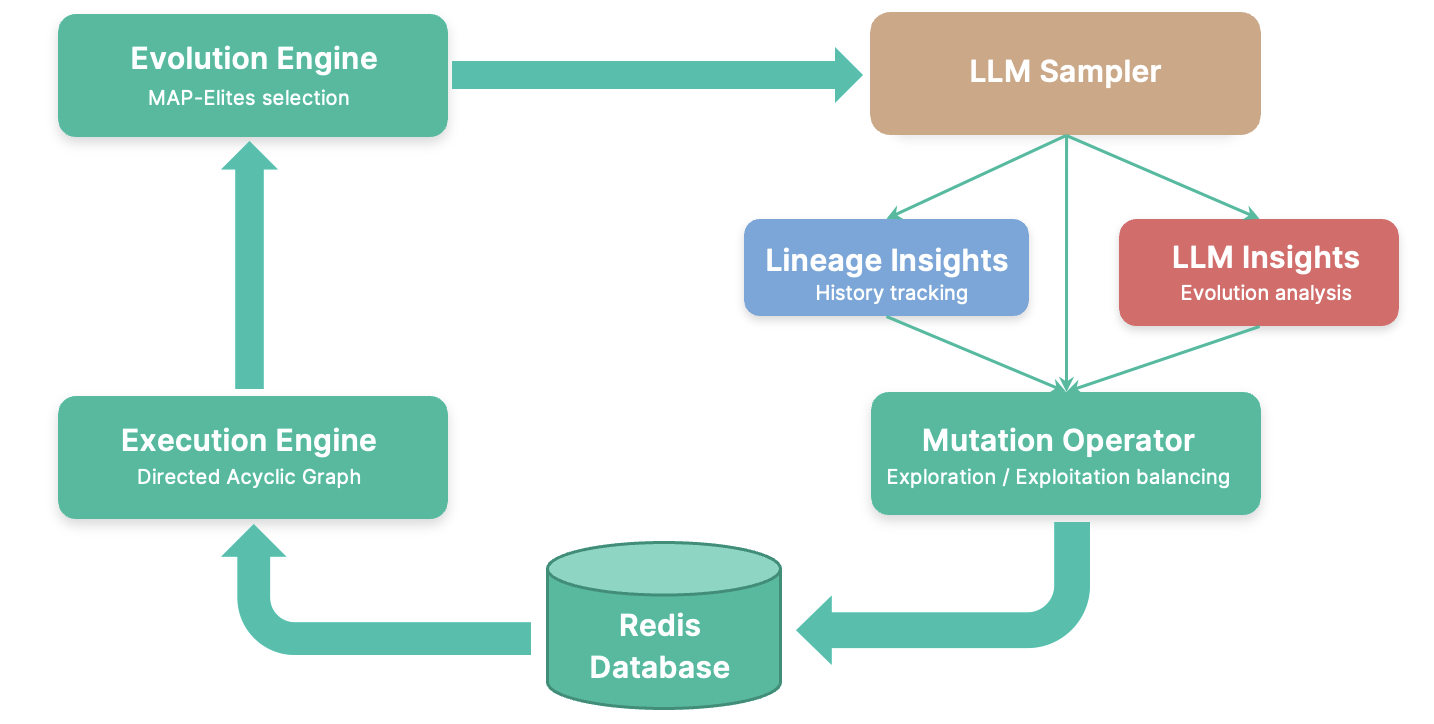}
    \caption{\textbf{GigaEvo system architecture.} 
    The framework comprises four core components (highlighted in green): (i) \textbf{Redis Database} --- a Redis storage for evolutionary units, which contain UUID, source code, lifecycle state, metrics, and lineage information, supporting concurrent access via optimistic concurrency control; (ii) \textbf{DAG Execution Engine} --- an asyncio-based pipeline processing programs through configurable stages (execution, validation, complexity analysis, LLM inference); (iii) \textbf{Evolution Engine} --- an asynchronous loop implementing MAP-Elites quality-diversity algorithm that maps programs to behavior space cells (fitness × validity), performs fitness-proportional selection of elites, and supports single-island and multi-island configurations with periodic migration; and (iv) \textbf{Mutation Operator} --- a LangGraph-based agent that constructs prompts from task descriptions, parent code, metrics, insights, and lineage analyses, then invokes LLMs to generate offspring programs using rewrite-based or diff-based mutation modes with multi-model routing support. All components interact through the Redis storage layer, enabling modular, concurrent evolutionary search.}
    \label{fig:scheme}
\end{figure}

The central abstraction of the framework is the \texttt{Program}, which represents an evolutionary unit.
Each \texttt{Program} contains a unique identifier (UUID), the program source code, a lifecycle state (e.g., \texttt{FRESH}, \texttt{RUNNING}, \texttt{COMPLETE}), evaluation metrics, genealogical lineage information, and the outputs of executed DAG stages (described below).
Programs are stored in a Redis-based storage layer with optimistic concurrency control via atomic counters.
This design supports concurrent evolutionary runs sharing a common program archive and enables efficient real-time lineage tracking.

\paragraph{DAG Execution Engine.}
To construct rich program contexts for LLM-based mutation, we implement a Directed Acyclic Graph (DAG) execution engine based on Python's \texttt{asyncio}.
The engine supports concurrent execution at two levels: multiple programs can be processed simultaneously, and within each program, independent stages execute in parallel.
Each stage is a self-contained operation implementing arbitrary logic—Python code execution, validation, complexity analysis, LLM inference, or agentic pipelines.
Stages are connected by two types of dependencies:
(i) \emph{data flow edges}, which pass outputs between stages (e.g., execution results flow into validation), and
(ii) \emph{execution-order dependencies}, which enforce sequential constraints without data transfer (e.g., ensuring metrics are computed before generating insights).
A \texttt{DAGAutomata} component handles scheduling, caching, and automatic stage skipping based on preconditions.
This design decouples program-processing logic from the execution engine, enabling flexible pipeline composition.
Following \textit{AlphaEvolve}, pipelines may employ cascading validation: lightweight checks filter failing programs early, while promising candidates undergo expensive evaluation. 
All processed programs and stage outputs are persisted to Redis for use by the evolutionary engine.

\paragraph{Evolutionary Engine.}
A dedicated asynchronous loop continuously monitors storage for newly completed programs and processes them through the evolutionary strategy.
Following \textit{AlphaEvolve}, we implement MAP-Elites (Multi-dimensional Archive of Phenotypic Elites), a quality-diversity algorithm that maintains a diverse archive of high-performing solutions.
The system supports both single-island and multi-island configurations.
In the single-island variant, programs are mapped to a two-dimensional behavior space discretized into cells based on: (i) fitness—the primary optimization metric with user-specified bounds, and (ii) validity—a binary indicator of syntactic and execution correctness.
Each program is mapped to its behavior cell; if the cell is empty or the program outperforms the current elite, it replaces the occupant and enters the \texttt{EVOLVING} state; otherwise, it is marked \texttt{DISCARDED}.
At each iteration, elites are sampled for mutation using fitness-proportional selection.
The multi-island extension maintains multiple independent MAP-Elites archives with distinct behavior spaces (e.g., fitness vs.\ complexity), enabling exploration of diverse trade-offs. 
Periodic migration exchanges top-performing programs between islands.

\paragraph{Mutation Operator.}
The mutation operator is implemented as a LangGraph-based agent that orchestrates prompt construction, LLM inference, and response parsing.
Given selected parent programs, the agent constructs a prompt incorporating the task description, parent code, metrics, generated insights, and lineage analyses (detailed below).
While \textit{AlphaEvolve} emphasizes generating \emph{diffs}, we found that many open-source models struggle to reliably produce syntactically correct diffs.
Our implementation supports both diff-based and rewrite-based mutation modes; the rewrite mode generates complete programs while carefully prompting the model to localize changes.
The system also supports multi-model routing, enabling heterogeneous LLM ensembles where different stages (e.g., insights vs.\ mutation) employ distinct models.
In practice, the rewrite strategy produces robust, syntactically valid programs with low failure rates.

\paragraph{Flexible configuration system.}
To facilitate rapid experimentation and reproducibility, the system employs Hydra \citep{Yadan2019Hydra} for hierarchical configuration management.
The configuration space is decomposed into orthogonal components: problem specification, evolutionary algorithm (single- vs.\ multi-island), LLM configuration (single model, multi-model ensembles, heterogeneous temperature profiles), DAG pipeline topology, and execution parameters (timeouts, parallelism, logging).
Each component resides in a separate YAML file; users compose experiments by selecting components via command-line overrides (e.g., \texttt{problem.name=heilbronn llm=multi\_model algorithm=multi\_island}) without editing code.
The system provides preset \emph{profiles} for common scenarios—for instance, a \texttt{base} profile optimizes for fast iteration during initial research.

\paragraph{Problem specification interface.}
The framework adopts a directory-based problem specification that enables rapid prototyping of new optimization tasks.
Each problem is defined by a self-contained directory containing four core components:
(i)~\texttt{task\_description.txt}, a natural language specification provided to the LLM during mutation;
(ii)~\texttt{metrics.yaml}, a declarative schema defining fitness metrics, bounds, precision, and significance thresholds;
(iii)~\texttt{validate.py}, a Python module implementing a \texttt{validate()} function that executes generated code and computes metrics; and
(iv)~\texttt{initial\_programs/}, a directory of seed programs that bootstrap the evolutionary process. Alternatively, the initial population can be provided via a Redis database, which is convenient when launching a new run from an already-obtained high-quality population. Certain problems require \emph{context}—pre-defined data to pass to the function being evolved, such as training datasets for machine learning problems. This is supported via DAG wiring by providing an input to the code evaluation node.

The metrics schema supports rich metric metadata: each metric declares whether higher values are better, numerical bounds for MAP-Elites discretization, decimal precision for display, and significance thresholds for detecting meaningful improvements.
A distinguished \texttt{is\_primary} flag identifies the optimization objective.
The validator function receives the output of the generated code's \texttt{entrypoint()}, performs problem-specific checks (e.g., constraint satisfaction, geometric validity), and returns a dictionary of metrics including a mandatory \texttt{is\_valid} indicator. If an exception occurs during execution, the error trace is provided as context to the insight generation stages (described below), which immediately provide feedback to the mutation operator on fixing these issues.

This modular design decouples problem definition from the execution engine: users add new problems by creating a directory with the required files—no modifications to core system code are necessary.
The \texttt{ProblemContext} and \texttt{ProblemLayout} classes provide standardized loading and access to problem assets, while the DAG pipeline automatically wires validation and metrics computation stages from the problem directory.
Initial programs may implement diverse strategies (e.g., constructive heuristics, random sampling, classical algorithms), providing the evolutionary process with a varied starting population.
This architecture enabled rapid prototyping and iteration on multiple problem variants during development.

\section{Experimental setup}
Following the approach in \textit{AlphaEvolve}, which leverages LLMs to generate guidance for constructing rich program contexts for mutation, we implement two complementary analysis systems: individual program insights and multi-stage bidirectional lineage tracking. We go into details of both systems below.

\paragraph{Insight generation.}
The \texttt{InsightsStage} analyzes individual programs by invoking an LLM with the task description, program code, metrics, and execution exceptions.
It produces structured insights categorized by type (e.g., algorithmic, structural), effect (beneficial, harmful, neutral), and severity (low, medium, high).
For instance, in a geometric optimization task, one generated insight reads:
\begin{quote}
\texttt{algorithmic [harmful] (high): Missing inter-point distance constraints. Implement minimum separation threshold to prevent degenerate triangles.}
\end{quote}

\paragraph{Lineage analysis.}
The lineage analysis subsystem consists of multiple coordinated stages.
The \texttt{LineageStage} analyzes each parent$\to$current program transition: it fetches the program's parent programs and invokes an LLM to produce structured transition analyses explaining code changes and metric shifts.
Two collector stages, \texttt{AncestorProgramIds} and \texttt{DescendantProgramIds}, select which ancestors and descendants merit detailed analysis (e.g., highest-fitness programs) using configurable selection strategies.
The \texttt{LineagesFromAncestors} stage filters the current program's parent$\to$current analyses to those from selected ancestors, while \texttt{LineagesToDescendants} retrieves current$\to$descendant analyses from descendants' stored \texttt{LineageStage} results.
This architecture enables bidirectional lineage tracking: the system gathers insights both from how the current program improved over selected ancestors and how selected descendants further improved upon it.
For example, a lineage analysis might read:
\begin{quote}
    \texttt{Lineage (imitation): +0.19 after clarifying output format → enhanced consistency improved model comprehension.}
\end{quote}

\begin{figure}[htb!]
    \centering
    \includegraphics[width=0.95\linewidth]{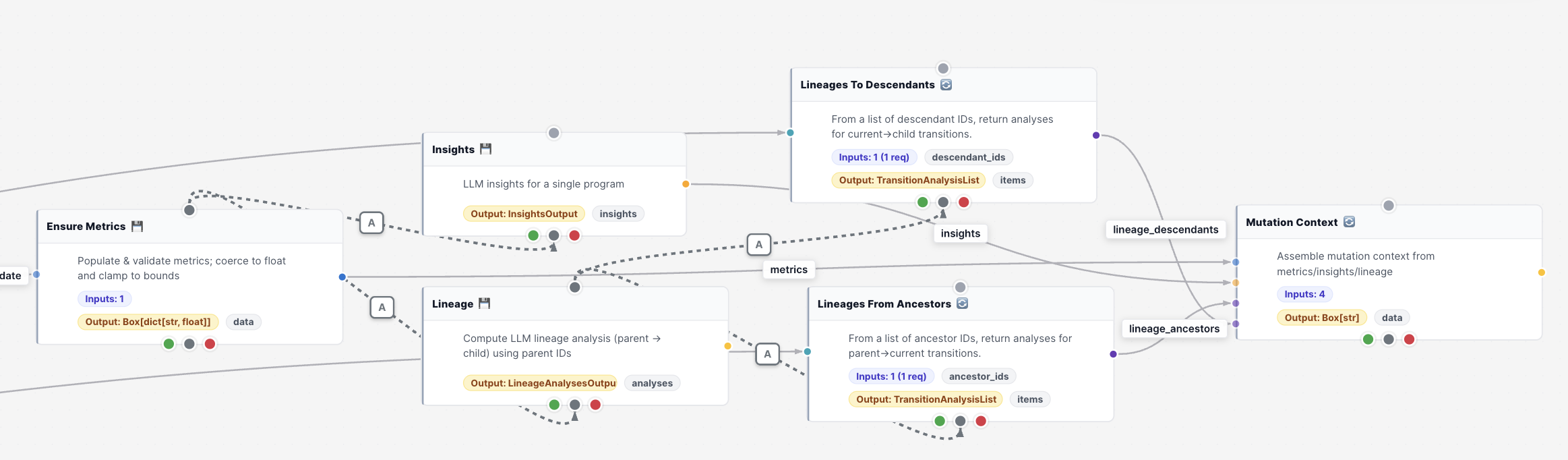}
    \caption{Part of the DAG pipeline in the GUI used to process programs in our experiments. 
    Programs are first validated (\texttt{ValidateCodeStage}), executed (\texttt{CallProgramFunction}), and evaluated via a problem-specific validator (\texttt{CallValidatorFunction}). 
    Complexity metrics (\texttt{ComputeComplexityStage}) are merged with validation metrics (\texttt{MergeMetricsStage}) and ensured complete via \texttt{EnsureMetricsStage}.
    Individual program insights are generated (\texttt{InsightsStage}), while lineage analysis proceeds through multiple stages: \texttt{LineageStage} computes parent$\to$current transition analyses; \texttt{AncestorProgramIds} and \texttt{DescendantProgramIds} select relevant relatives; \texttt{LineagesFromAncestors} and \texttt{LineagesToDescendants} gather bidirectional lineage insights.
    Finally, \texttt{MutationContextStage} assembles all information into a structured context for the mutation operator. 
    The framework supports arbitrary DAG topologies through declarative configuration.}
    \label{fig:dag}
\end{figure}

These stages are incorporated into the DAG pipeline shown in \cref{fig:dag}.

\paragraph{Mutation context assembly.}
The \texttt{MutationContextStage} aggregates outputs from upstream stages—code, metrics, insights, and bidirectional lineage analyses (from ancestors and to descendants)—into a structured dictionary stored in program metadata.
This context is retrieved during mutation to provide the LLM with comprehensive information about parent programs, including what strategies succeeded in the past and which improvements descendants achieved.

\section{Benchmark problems}
We evaluate \emph{GigaEvo} on four challenging optimization problems studied in~\cite{novikov2025alphaevolve,georgiev2025mathematical} to assess the reproducibility of published results and validate our implementation.
These problems span computational geometry, discrete optimization, and high-dimensional sphere packing, each requiring the discovery of mathematical constructions with extremal properties.
For each problem, we describe the mathematical formulation, the evaluation approach used in the original work, and the target benchmarks we aim to reproduce. For thorough descriptions and further background, we refer the reader to \cite{georgiev2025mathematical}; here we briefly discuss important details. For simplicity, all experiments use a \emph{single-island} setup. Our preliminary experiments demonstrated no apparent benefit from the multi-island setup, which only increased system complexity. As the backbone LLM, we use \textit{Qwen3-235B-A22B-Thinking-2507}~\citep{qwen3technicalreport} with default settings.

\subsection{Heilbronn triangle problem.}
\begin{figure}[htb!]
    \centering
    \includegraphics[width=0.95\linewidth]{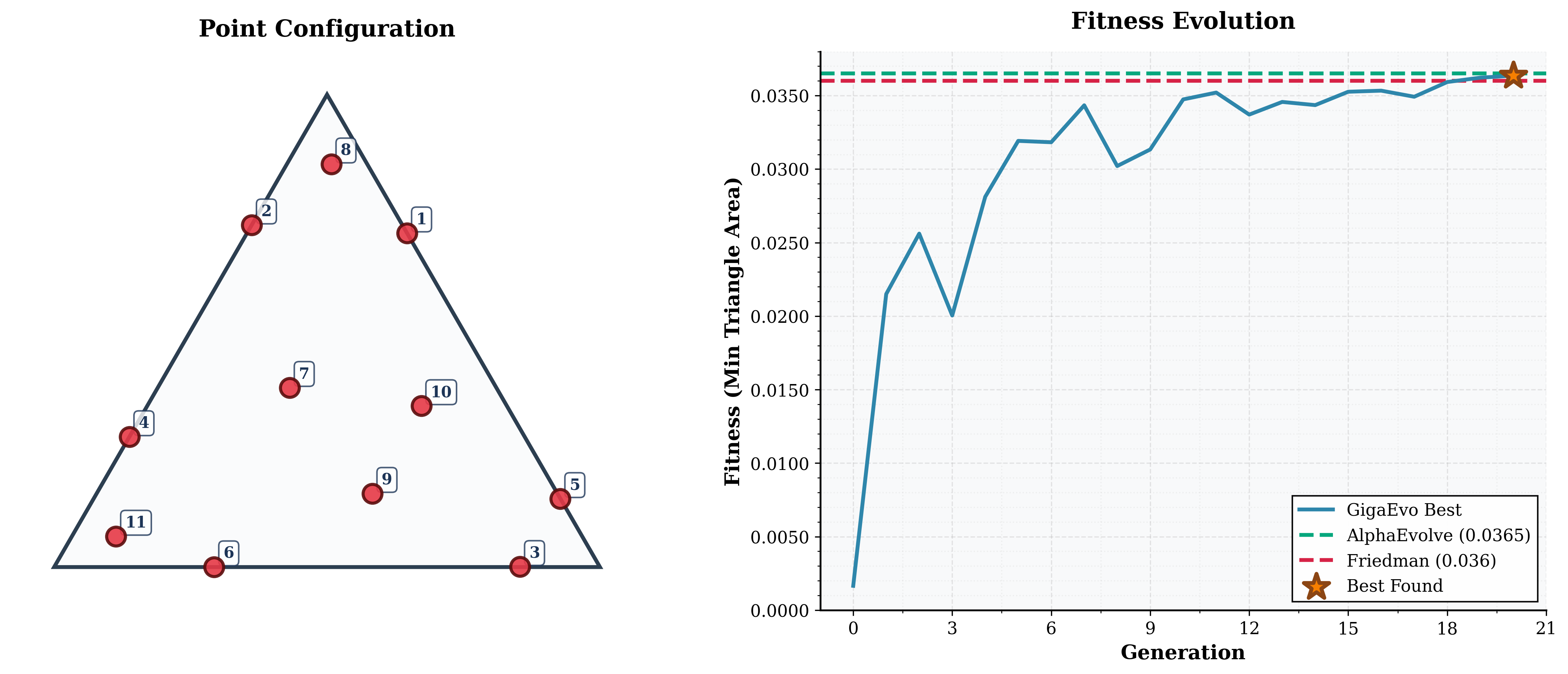}
    \caption{\textbf{Heilbronn triangle problem results for $n=11$ points.}
        \textbf{Left:} Point configuration discovered by GigaEvo, maximizing the minimum triangle area formed by any three points in a unit-area equilateral triangle.
        The evolved program places points both on the boundary and in the interior, exploiting geometric symmetries while avoiding near-collinear triplets.
        \textbf{Right:} Fitness evolution over 20 evolution generations, showing rapid initial progress followed by refinement.
        The dashed lines indicate benchmark results: AlphaEvolve achieved $0.0365$ (green), while the previous best \citep{friedman_main_website} was approximately $0.036$ (red).
        GigaEvo achieved a fitness of $0.0364$, which is $10^{-4}$ below the reported AlphaEvolve result.
        By visual inspection (fig. 26 left panel in \cite{georgiev2025mathematical}), the point configurations are nearly identical, suggesting GigaEvo successfully reproduced the AlphaEvolve solution.
        The small fitness difference could be eliminated with additional computational budget for fine-tuning the simulated annealing parameters in the final optimization phase. This demonstrates that our implementation captures the essential algorithmic principles underlying AlphaEvolve's success.}
    \label{fig:triangle}
\end{figure}

For any $n \geq 3$ and any convex body $K$ in the plane, let $C(n,K)$ be the largest quantity such that in every configuration of $n$ points in $K$, there exists a triple of points determining a triangle of area at most $C(n,K)$ times the area of $K$.
This problem, posed by Heilbronn in the 1950s, asks to establish upper and lower bounds on $C(n,K)$. Following the discussed paper, we consider the case $n = 11$ with $K$ being the equilateral triangle of unit area.
The function to be evolved proposes point configurations, and the score is the area of the smallest triangle formed by any three points. If constraints are violated, an exception is raised.
The target benchmark is the result from \cite{novikov2025alphaevolve}, which achieved a minimum triangle area of $0.0365$ (improving upon the previous best of approximately $0.036$ reported in~\cite{friedman_main_website}).
The optimization landscape is highly non-convex: small coordinate perturbations can drastically reduce fitness due to near-collinearities or clustering, making this problem particularly challenging for evolutionary search.
\paragraph{Results.}
Our system successfully recovered a solution nearly identical to that found by \emph{AlphaEvolve}, as shown in \cref{fig:triangle}. The obtained metric value was slightly below in the 4th decimal digit, but in such cases, numerical parameters can be fine-tuned using an additional stage such as CMA-ES to marginally improve the metric. By visual inspection of the configuration in \emph{AlphaEvolve}, we confirmed that the same arrangement was discovered. The task description provided to the mutation operator and the obtained program are available in Appendix~\ref{sec:heilbron}.

\paragraph{Circle packing in a square.}
For any $n \geq 1$, let $C(n)$ denote the largest sum $\sum_{i=1}^n r_i$ of radii such that one can place $n$ disjoint open disks of radii $r_1,\dots,r_n$ inside the unit square.
Existing upper bounds on this quantity may be found at~\cite{friedman_main_website}.
We aim to reproduce the improvements reported in \cite{novikov2025alphaevolve}, which discovered new constructions advancing these bounds.

\begin{figure}[htb!]
    \centering
    \includegraphics[width=0.4\linewidth]{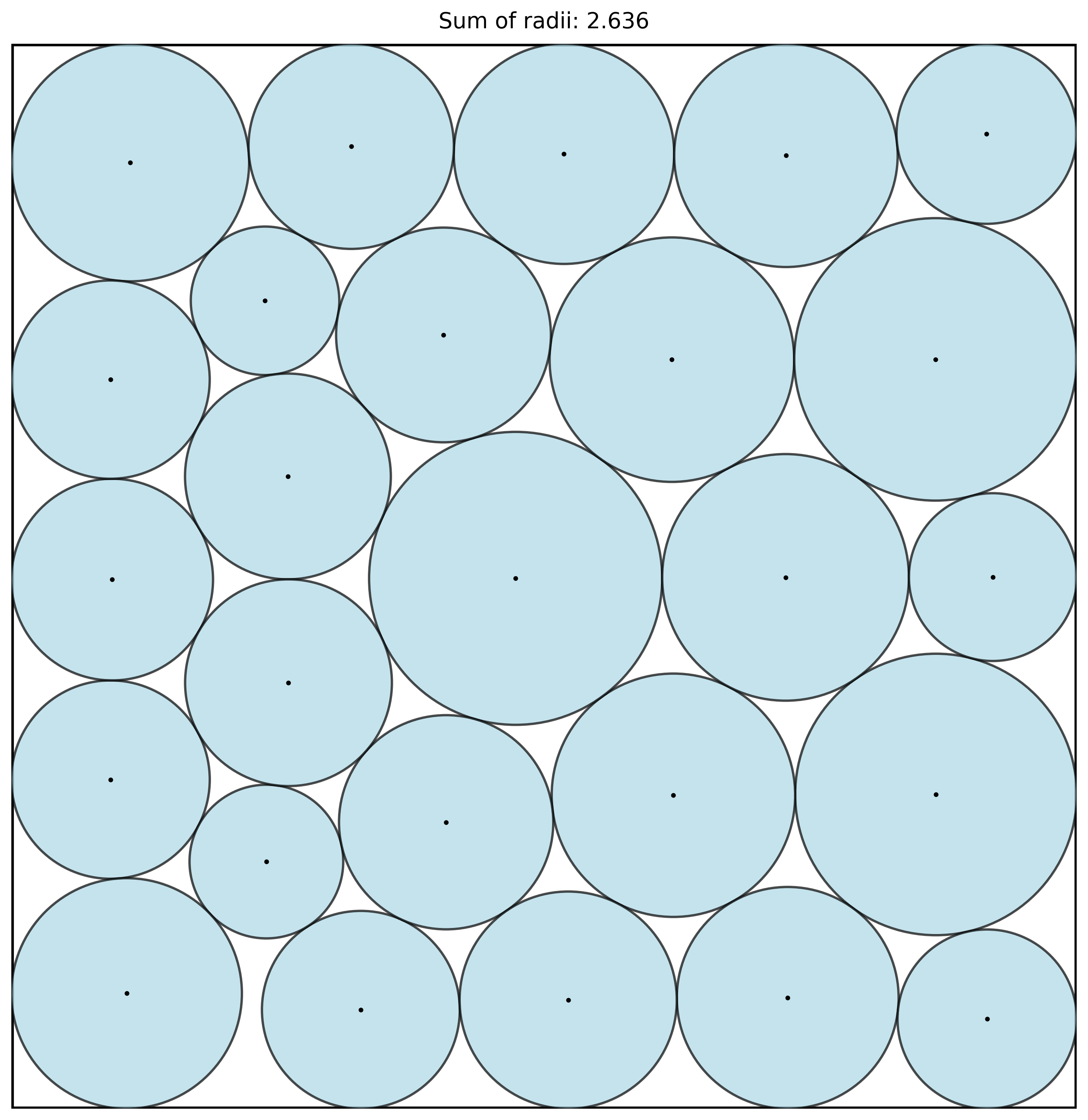}
    \caption{\textbf{Circle packing in unit square for $n=26$ circles.}
        Optimal configuration discovered by GigaEvo, maximizing the sum of radii $\sum_{i=1}^{26} r_i$ of disjoint circles packed within the unit square.
        GigaEvo achieved a total radius sum of $2.636$, slightly surpassing the AlphaEvolve result of $2.635$ reported in~\cite{novikov2025alphaevolve}.
        Visual comparison with the AlphaEvolve solution (fig. 28 in \cite{georgiev2025mathematical}) reveals nearly identical qualitative structure, with minor improvements likely attributable to refined local optimization of centers and radii.
        This demonstrates that our implementation successfully reproduces and marginally improves upon AlphaEvolve's results through effective exploration of the constrained geometric optimization landscape.}
    \label{fig:circle}
\end{figure}

Unlike fixed-radius circle packing, this variable-radius formulation allows the optimizer to explore heterogeneous disk configurations, balancing large disks in open regions against small disks filling interstices.
This characteristic makes it particularly suitable for evolutionary search, where incremental improvements emerge through mutation and selection pressure.
The system must jointly optimize disk positions and radii while maintaining non-overlap constraints and boundary compliance—a constrained nonlinear optimization problem with complex geometric interactions. We consider $n=26$ and $n=32$.
\paragraph{Results.} We extensively experimented with this task during the preliminary stage of developing our pipeline. Our system achieved and slightly surpassed the reported AlphaEvolve value for $n=26$; namely, we obtained $2.63598$ (compared to the reported value $2.635$). A visual representation of this configuration is shown in \cref{fig:circle}. By visual inspection, we observe that this configuration is nearly qualitatively identical to the AlphaEvolve result, with marginal improvements likely due to refined optimization of centers and radii. For $n=32$, our framework achieved a score of $2.939$, which is noticeably better than the previous state-of-the-art of $2.937$.

\subsection{Kissing numbers in high dimensions.}
For a dimension $n \geq 1$, define the \emph{kissing number} $C(n)$ to be the maximum number of non-overlapping unit spheres that can be arranged to simultaneously touch a central unit sphere in $n$-dimensional space.
This problem has been studied since at least 1694, when Isaac Newton and David Gregory discussed what $C(3)$ would be.
The cases $C(1) = 2$ and $C(2) = 6$ are trivial.
The four-dimensional problem was solved by ~\cite{musin2008kissing}, who proved that $C(4)=24$.
In dimensions 8 and 24, the problem is also solved: the $E_8$ lattice and the Leech lattice give kissing numbers of $C(8)=240$ and $C(24) = 196560$, respectively. In the report \cite{novikov2025alphaevolve}, an improved lower bound for $C(11)$ was obtained, raising it from 592 to 593. In \cite{georgiev2025mathematical}, multiple dimensionalities were studied, confirming known lower bounds.

We adopt the integer lattice formulation employed in \emph{AlphaEvolve}.
Lower bounds on $C(n)$ are established by constructing explicit configurations of integer vectors satisfying geometric constraints.
The task is to find distinct non-zero integer vectors $\{x_i\} \subset \mathbb{Z}^n$ such that:
(i)~\emph{single shell}: all vectors lie on a common shell, $\|x_i\|^2 = r^2$ for some integer $r^2 > 0$;
(ii)~\emph{separation certificate}: pairwise squared distances satisfy $\min_{i \neq j} \|x_i - x_j\|^2 \geq r^2$.
After appropriate scaling by $t = 2/\sqrt{r^2}$, these vectors yield a valid kissing configuration with all sphere centers at radius 2 and pairwise distances at least 2.

The fitness metric is the cardinality $N$ of the constructed set.
Validation employs exact integer arithmetic: squared norms are computed as $\sum_{k=1}^n x_{i,k}^2$ using Python's arbitrary-precision integers, and pairwise squared distances $\|x_i - x_j\|^2$ are checked exhaustively. The evolved programs must discover constructive algebraic methods—such as exploiting lattice symmetries, orbit decompositions, or algebraic number field constructions—rather than combinatorial enumeration, which becomes computationally infeasible at target scales.

We target dimension 12 with the goal of improving the established lower bound of $N \geq 840$. 
\paragraph{Results.} We found that our default model \emph{Qwen3-235B-A22B-Thinking-2507} struggled to make progress on this problem, plateauing around fitness $\sim 500$. We replaced Qwen with Google Gemini-2.5-Flash \cite{comanici2025gemini}, which readily achieved values of $840$ (the established lower bound). However, we were unable to improve beyond this value. This mirrors the findings in \cite{georgiev2025mathematical}, where for most dimensions, the known lower bound was recovered but no improvements were achieved. We speculate that this construction is well-known and was present in the Gemini pre-training corpora. Overall, we believe the kissing number problem is one of the most challenging problems and can serve as a future benchmark.

\subsection{Bin Packing}

Beyond the mathematical optimization problems studied in AlphaEvolve, we evaluate GigaEvo on the one-dimensional online bin packing problem to assess generalization to algorithmic design tasks.
In this problem, items of varying sizes arrive sequentially and must be irrevocably assigned to bins of fixed uniform capacity, with the objective of minimizing the total number of bins used.

\paragraph{Problem formulation.}
The task is to evolve packing heuristics that decide, for each incoming item, which bin to place it in (or whether to open a new bin).
Item sizes are sampled from stochastic distributions: either uniform or Weibull.
Performance is measured by comparing the number of bins used by the online heuristic against the optimal offline solution (computed with full knowledge of the sequence).
The evaluation metric is \emph{excess bins}: the difference between bins used by the heuristic and the optimal number.
Lower excess indicates better performance.

\paragraph{Experimental setup.}
We employed Llama 3~\citep{grattafiori2024llama} models in two configurations: the 7B parameter version for rapid exploration and the 70B parameter version for refinement.
The evolutionary process required approximately 200 LLM calls.
An ablation study revealed that the smaller 7B model drove exploration, while the 70B model primarily accelerated convergence rather than generating qualitatively novel strategies.
We determined an optimal temperature of 0.6: lower temperatures increased syntactic validity above 50\% but reduced solution diversity, hindering evolutionary progress.

\paragraph{Results.}
GigaEvo successfully replicated the state-of-the-art result achieved by FunSearch \citep{romera2024mathematical} for the uniform distribution.
More significantly, we established a \emph{new state-of-the-art for the Weibull distribution}, reducing excess bin usage from 0.68\% (FunSearch) to 0.55\%.
This demonstrates that GigaEvo generalizes beyond geometric optimization to algorithmic discovery tasks and can discover novel heuristics competitive with specialized prior work.

\subsection{Prompt and Agent Evolution}

We additionally evaluate GigaEvo on evolving solutions that leverage downstream LLMs for natural language processing tasks. The evolutionary process targets two distinct solution architectures: \textit{prompts}, which provide direct instructions for task completion in a single LLM call, and \textit{agents}, which implement multi-step reasoning strategies that combine multiple LLM calls to construct a solution. This setting tests whether evolutionary optimization can discover effective prompt engineering and feature integration strategies for real-world applications.

\paragraph{Problem Formulation.}

We evaluate GigaEvo on \textsc{The Jigsaw - Agile Community Rules Classification}\footnote{\url{https://www.kaggle.com/competitions/jigsaw-agile-community-rules}} competition hosted on the \textsc{Kaggle} platform. Given a Reddit comment $c$ and a subreddit rule $r$, the task requires learning a function $f: (c, r) \rightarrow \{0, 1\}$ that maps comment-rule pairs to binary labels, where 1 indicates a rule violation.

The dataset comprises $2048$ training examples and approximately $20$ times more test examples, with each instance containing the comment body, rule text, subreddit identifier, and few-shot demonstrations (two positive and two negative examples). Performance is evaluated using the per-rule Area Under the ROC Curve (AUC), averaged across all test rules. The test set includes $4$ hidden rules unseen during training, challenging solution to generalize to novel community norms.

\paragraph{Experimental Setup.}

We used \textsc{Qwen2.5-14B-Instruct}~\cite{qwen2.5} as the base model to perform the classification task, with \textsc{gpt-oss-120b}~\cite{openai2025gptoss120bgptoss20bmodel} (high reasoning effort) serving as the evolutionary engine. The evolutionary process lasted approximately $60$ generations, with each prompt and agent candidate evaluated on the full training dataset. 

Initial experiments revealed that both evolved prompts and agents consistently attempted to construct custom similarity measures between comments and the provided metadata. To assist the evolutionary engine, we used \textsc{Qwen3-Embedding-0.6B}~\cite{qwen3embedding} and computed: (1) cosine similarity between the rule and comment, (2) average cosine similarity between the comment and positive examples, and (3) average cosine similarity between the comment and negative examples. These similarity scores were provided as additional input features to all candidate solutions.

\paragraph{Results.}

The baseline prompt achieved AUC of $0.673$ on the training data and $0.670$ on the test data. After $60$ generations, GigaEvo was able to evolve the prompt that demonstrated substantial improvement, scoring $0.794$ AUC on the training data and $0.783$ on the test data, representing a $11.3$ percentage points AUC gain on the test set. Incorporating an agent-based approach further yielded additional gains, achieving $0.811$ AUC on training data and $0.803$ on test data. Crucially, this agent remained computationally efficient by requiring only a single LLM call, similar to the evolved prompt configuration. The performance improvement was obtained by ensembling the LLM's predicted probability with the precomputed Qwen embedding similarities through a calibrated sigmoid function. These experiments demonstrate that GigaEvo can discover effective architectures for integrating neural predictions with traditional feature-based methods.

\section{Conclusions}

In this report we presented GigaEvo, an open-source framework for LLM-driven evolutionary computation. Our system provides modular implementations of key components: MAP-Elites quality-diversity algorithms, asynchronous DAG-based evaluation pipelines, LLM mutation operators with bidirectional lineage tracking, and flexible configuration management, which enables reproducibility studies and rapid prototyping of new problems.

\paragraph{Implementation insights.}
Several design decisions proved critical:
(i) Rewrite-based mutation with careful prompting for localized changes provides more robust program generation than diff-based approaches for open-source models;
(ii) Bidirectional lineage tracking (analyzing both ancestor innovations and descendant improvements) enriches mutation context;
(iii) Heterogeneous LLM routing enables leveraging different models' strengths (e.g., Qwen for geometry, Gemini for discrete optimization);
(iv) Declarative Hydra-based configuration substantially reduces iteration time during problem development.

\paragraph{Limitations and future work.}
Our single-island experiments showed no clear benefit of multi-island MAP-Elites, contrary to expectations; further investigation of behavior space design is warranted.
For geometric problems requiring fine-grained numerical optimization, hybrid approaches integrating continuous local search with LLM-driven structural innovation may be beneficial.
The framework currently focuses on single-file Python programs; extending to multi-file projects and other programming languages remains future work.

\paragraph{Broader impact.}
By providing a well-documented, modular implementation of LLM-evolutionary methods, GigaEvo aims to lower the barrier to entry for researchers investigating this emerging paradigm.
The reproducibility of AlphaEvolve results with our alternative architecture suggests that the core principles—quality-diversity search, rich program contexts via lineage analysis, and LLM-driven mutation—are robust to implementation variations.
We hope GigaEvo serves as a foundation for further research in automated algorithm discovery and mathematical construction.

The framework and all experimental code are available as open source at \href{https://github.com/AIRI-Institute/gigaevo-core}{https://github.com/AIRI-Institute/gigaevo-core}.

\bibliography{iclr2024_conference}
\bibliographystyle{iclr2024_conference}

\newpage
\appendix

\section{Heilbronn Triangle Problem: Task Description and Evolved Solution}{
\label{sec:heilbron}
}

\subsection{Task Description}

The following task description was provided to the LLM during the evolutionary process:

\begin{quote}
\textbf{TASK DEFINITION -- HEILBRONN TRIANGLE PROBLEM (11 POINTS)}

\textbf{Challenge:} High-dimensional non-convex geometric optimization. Place 11 distinct points inside a unit-area equilateral triangle to maximize the smallest triangle area formed by any three points. Target: \texttt{min\_area} $\geq$ 0.0365

\textbf{OBJECTIVE}

Return 11 distinct 2D coordinates inside a unit-area equilateral triangle such that:
\begin{enumerate}
    \item \textbf{(Containment)} All points lie inside or on the boundary of the enclosing triangle
    \item \textbf{(Distinctness)} All points are distinct (no duplicates)
    \item \textbf{(Non-degeneracy)} No three points are collinear
    \item \textbf{(Maximality)} The minimum area among all $\binom{11}{3} = 165$ triangles is maximized
\end{enumerate}

\textbf{Output:} (11, 2) NumPy array $|$ \textbf{Fitness:} \texttt{min\_triangle\_area} $|$ \textbf{Goal:} \texttt{min\_triangle\_area} $\geq$ 0.0365

\textbf{CONSTRAINTS}
\begin{itemize}
    \item Enclosing triangle: equilateral, area = 1.0, flat-bottomed, base along x-axis, apex at (0, 0)
    \item Exactly 11 distinct points, all within or on boundary
    \item No collinear triplets (all 165 triangles must have non-zero area)
\end{itemize}

\textbf{FAILURE MODES}
\begin{itemize}
    \item Point clustering (small triangle areas); nearly collinear triplets (flat triangles)
    \item Duplicate or overlapping coordinates; boundary violations (points outside triangle)
    \item Incorrect enclosing triangle geometry (wrong area, orientation, or vertices)
    \item Float precision errors causing false collinearity detection
\end{itemize}

\textbf{HELPER FUNCTIONS}
\begin{itemize}
    \item \texttt{get\_unit\_triangle()} $\rightarrow$ tuple of 3 arrays (A, B, C), each shape (2,) - vertices of unit-area triangle
    \item \texttt{get\_smallest\_triangle\_area(coordinates)} $\rightarrow$ float - minimum area among all $\binom{n}{3}$ triangles
    \item \texttt{is\_inside\_triangle(points, a, b, c)} $\rightarrow$ bool - True if all points inside/on triangle boundary
\end{itemize}

\textbf{OUTPUT FORMAT:}

Implement \texttt{def entrypoint():} that returns a (11, 2) NumPy array with all points inside the unit-area equilateral triangle (flat-bottomed, apex at origin). Fix random seeds if using randomness (e.g., \texttt{np.random.seed(42)}).
\end{quote}

\newpage
\subsection{Evolved Solution Code}

The following program was evolved by GigaEvo (Program ID: 64330a3d-bcf4-44e0-ad0d-96337addba30, Fitness: 0.036354, Generation: 20). This solution employs sophisticated techniques including Halton sequences for quasi-random sampling, adaptive simulated annealing, critical triangle resolution, and fitness-adaptive boundary distribution.

{\footnotesize
\begin{verbatim}
# Best Heilbronn Triangle Program
# Generated by GigaEvo
# Program ID: 64330a3d-bcf4-44e0-ad0d-96337addba30
# Fitness: 0.036354, Generation: 20

import numpy as np
from helper import get_unit_triangle, get_smallest_triangle_area, is_inside_triangle

def entrypoint() -> np.ndarray:
    np.random.seed(42)
    
    def halton(index, base):
        result = 0.0
        f = 1.0 / base
        i = index
        while i > 0:
            result += f * (i % base)
            i = i // base
            f = f / base
        return result

    A, B, C = get_unit_triangle()
    L = np.linalg.norm(B - A)
    threshold_area = 1e-6
    num_restarts = 25
    best_points = None
    best_min_area = -1.0
    
    # Calculate edge vectors and perpendiculars for boundary perturbations
    AB = B - A
    BC = C - B
    CA = A - C
    perp_AB = np.array([-AB[1], AB[0]]) / np.linalg.norm(AB)
    perp_BC = np.array([-BC[1], BC[0]]) / np.linalg.norm(BC)
    perp_CA = np.array([-CA[1], CA[0]]) / np.linalg.norm(CA)
    
    def get_base_pairs(restart, best_local_min):
        # Geometrically adaptive threshold based on fitness progress
        dynamic_threshold = 0.032 + 0.004 * min(best_local_min / 0.0365, 1.0)
        if best_local_min > dynamic_threshold:
            return [(2, 5), (5, 2), (3, 7), (7, 3), (5, 7), (7, 5), (3, 5), (5, 3)]
        elif restart % 4 == 0:
            return [(2, 3), (3, 2), (2, 7), (7, 2), (3, 5), (5, 3)]
        else:
            return [(2, 5), (5, 2), (3, 5), (5, 3), (7, 2), (2, 7)]
    
    # Fitness-adaptive boundary distribution
    def create_boundary_points():
        points = []
        boundary_ratio = 0.65 + 0.15 * (best_min_area / 0.0365)
        total_boundary = max(5, min(8, int(11 * boundary_ratio)))
        
        ab_count = max(2, int(total_boundary * 0.34))
        bc_count = max(2, int(total_boundary * 0.33))
        ca_count = total_boundary - ab_count - bc_count
        
        for i in range(ab_count):
            t = (i + 0.5) / (ab_count + 0.5)
            P = A + t * (B - A)
            perturbation = 0.003 * L * np.sqrt(max(best_min_area, 0.001))
            P += np.random.uniform(-perturbation, perturbation) * perp_AB
            points.append(P)
        
        for i in range(bc_count):
            t = (i + 0.5) / (bc_count + 0.5)
            P = B + t * (C - B)
            perturbation = 0.003 * L * np.sqrt(max(best_min_area, 0.001))
            P += np.random.uniform(-perturbation, perturbation) * perp_BC
            points.append(P)
        
        for i in range(ca_count):
            t = (i + 0.5) / (ca_count + 0.5)
            P = C + t * (A - C)
            perturbation = 0.003 * L * np.sqrt(max(best_min_area, 0.001))
            P += np.random.uniform(-perturbation, perturbation) * perp_CA
            points.append(P)
        
        return np.array(points)

    for restart in range(num_restarts):
        np.random.seed(42 + np.random.randint(0, 1000))
        offset = np.random.randint(0, 1000000)
        base_pairs = get_base_pairs(restart, best_min_area)
        base1, base2 = base_pairs[np.random.randint(0, len(base_pairs))]
        points = []
        initial_dup_threshold = 0.005 * L

        # Enhanced initialization with fitness-adaptive boundary distribution
        if restart % 5 == 0 and best_min_area < 0.035:
            points = create_boundary_points()
            current_min = get_smallest_triangle_area(points)
        else:
            for i in range(3):
                found = False
                for attempt in range(1000):
                    idx_halton = offset + i * 1000 + attempt
                    r1 = halton(idx_halton, base1)
                    r2 = halton(idx_halton, base2)
                    u = np.sqrt(r1)
                    v = r2
                    P = (1 - u) * A + u * (1 - v) * B + u * v * C
                    
                    if i == 0:
                        points.append(P)
                        found = True
                        break
                    elif i == 1:
                        if np.linalg.norm(P - points[0]) > initial_dup_threshold * 2:
                            points.append(P)
                            found = True
                            break
                    else:
                        new_points_arr = np.array([points[0], points[1], P])
                        min_area = get_smallest_triangle_area(new_points_arr)
                        if min_area > threshold_area * 5:
                            points.append(P)
                            found = True
                            break
                if not found:
                    points = None
                    break
            
            if points is None or len(points) < 3:
                continue
            points = np.array(points)
            current_min = get_smallest_triangle_area(points)
        
        dup_threshold = max(0.002 * L, 0.08 * np.sqrt(current_min))
        
        # Simulated annealing for point insertion with critical triangle resolution
        # ... (additional optimization steps - code continues with SA iterations,
        # critical triangle resolution, and final refinement)
        
    return best_points
\end{verbatim}
}

\textit{Note: The full code is 443 lines. This excerpt shows the key algorithmic components. The complete implementation includes critical triangle resolution (lines 180-309), adaptive simulated annealing (lines 310-346, 359-400), and sophisticated fallback strategies.}

\section{Prompt And Agent Evolution Results}
\label{sec:prompt_nd_agent_appx}

\subsection{Final Prompt}

The following prompt was evolved by GigaEvo. This solution employs content step-by-step reasoning, analysis of multiple relationships between different metadata, incorporating similarities for decision, and final score calculation.

{\footnotesize
\begin{verbatim}
PROMPT_TEMPLATE: str = """You are a moderation assistant. 
Estimate the probability that the following Reddit comment violates the given rule.

Comment:
{body}

Rule:
{rule}

Similarity scores (higher = more similar):
- Rule‑comment similarity: {rule_sim}
- Similarity to violating examples: {avg_positive_sim}
- Similarity to non‑violating examples: {avg_negative_sim}

Violating examples:
- {positive_example_1}
- {positive_example_2}

Non‑violating examples:
- {negative_example_1}
- {negative_example_2}

Reasoning (think step‑by‑step):
1. Summarize the comment and identify any part that may relate to the rule.
2. Restate the rule in your own words.
3. Interpret each similarity score qualitatively: high (>=0.7), moderate (0.4‑0.7), low (<=0.4).
4. Compare the comment with the violating and non‑violating examples, 
noting similarities or differences.
5. Weigh the observations, giving the most importance to the rule‑comment similarity, 
then to the positive example similarity, 
and reducing the score for high negative‑example similarity.
6. Assign a probability of violation between 0.00 and 1.00, rounded to two decimal places.

Output only the final line in this exact format:
Answer: X.XX
"""
\end{verbatim}
}

\subsection{Final Agent}

The following agent was evolved by GigaEvo. This solution employs heuristic probability estimation using Qwen embedding similarities through a calibrated sigmoid function, estimating the probability using step-by-step reasoning with LLM in a single call, and then ensembling all probabilities to the final prediction. Each step uses optional fallbacks if the extraction fails.

{\footnotesize
\begin{verbatim}
import re
import math

class Agent:
    def __init__(self, client):
        self.client = client

        # Primary prompt
        self.primary_prompt = """{Similar to 'Final Prompt'}"""

    async def __call__(self, sample: dict[str, str]) -> float:
        # ------------------------------------------------------------------
        # 0. Parse similarity fields safely
        # ------------------------------------------------------------------
        try:
            rule_sim = float(sample.get("rule_sim", 0.0))
            avg_pos = float(sample.get("avg_positive_sim", 0.0))
            avg_neg = float(sample.get("avg_negative_sim", 0.0))
        except (ValueError, TypeError):
            rule_sim = avg_pos = avg_neg = 0.0

        # ------------------------------------------------------------------
        # 1. Heuristic prior (always available)
        # ------------------------------------------------------------------
        heuristic = self._heuristic(rule_sim, avg_pos, avg_neg)

        # ------------------------------------------------------------------
        # 2. Primary LLM call (moderate temperature for nuanced output)
        # ------------------------------------------------------------------
        prompt1 = self.primary_prompt.format(
            body=sample.get("body", ""),
            rule=sample.get("rule", ""),
            subreddit=sample.get("subreddit", ""),
            rule_sim=rule_sim,
            avg_positive_sim=avg_pos,
            avg_negative_sim=avg_neg,
            positive_example_1=sample.get("positive_example_1", ""),
            positive_example_2=sample.get("positive_example_2", ""),
            negative_example_1=sample.get("negative_example_1", ""),
            negative_example_2=sample.get("negative_example_2", ""),
        )
        resp1 = await self.client(
            prompt=prompt1,
            max_tokens=200,
            temperature=0.4,
        )
        prob_primary = self._extract_probability(resp1)

        # Fallback to heuristic if extraction fails
        if prob_primary is None:
            prob_primary = heuristic

        # 3. Weighted aggregation with normalization
        # ------------------------------------------------------------------
        # Updated weight scheme: primary 0.70, heuristic 0.30
        w_primary = 0.70
        w_heuristic = 0.30

        total = w_primary + w_heuristic
        final = (w_primary * prob_primary + w_heuristic * heuristic) / total

        # Clamp to valid probability range
        final = max(0.0, min(1.0, final))
        return final

    def _extract_probability(self, text: str):
        """
        Extract a probability from LLM output.
        Looks for a line starting with 'Answer:'.
        """
        m = re.search(r'Answer:\s*([-+]?\d*\.?\d+(?:[eE][-+]?\d+)?)', text, re.IGNORECASE)
        if m:
            try:
                val = float(m.group(1))
                # Convert percentage >1 to probability
                if val > 1.0:
                    val = val / 100.0
                return max(0.0, min(1.0, val))
            except ValueError:
                pass
        return None

    def _heuristic(self, rule_sim: float, avg_pos: float, avg_neg: float) -> float:
        """
        Similarity‑driven prior using a calibrated sigmoid.
        """
        # Linear combination emphasizing the gap between positive/negative examples
        x = 5.0 * (avg_pos - avg_neg) + 2.0 * rule_sim
        try:
            prob = 1.0 / (1.0 + math.exp(-x))
        except OverflowError:
            prob = 0.0 if x < 0 else 1.0
        return max(0.0, min(1.0, prob))
\end{verbatim}
}

\end{document}